\documentclass[sigconf]{acmart}

\usepackage{amssymb}
\usepackage{cleveref}
\usepackage{algorithm}
\usepackage{algpseudocode}  
\usepackage[disable]{todonotes}
\usepackage{subcaption} 
\usepackage{lipsum}
\usepackage{ragged2e}
\AtBeginDocument{%
  }

\setcopyright{acmlicensed}
\copyrightyear{2018}
\acmYear{2018}
\acmDOI{XXXXXXX.XXXXXXX}
\renewcommand\footnotetextcopyrightpermission[1]{}
\acmConference[Conference acronym 'XX]{Make sure to enter the correct
  conference title from your rights confirmation email}{June 03--05,
  2018}{Woodstock, NY}
\acmISBN{978-1-4503-XXXX-X/2018/06}

\newcommand{\ourwork}{\textsc{Galaxy}DiT}
\begin{document}

\title{\textsc{Galaxy}DiT: Efficient Video Generation with \underline{G}uidance \underline{Al}ignment and \underline{A}daptive Pro\underline{xy} in \underline{Di}ffusion \underline{T}ransformers}






\author{
Zhiye Song\textsuperscript{1},
Steve Dai\textsuperscript{2}, 
Ben Keller\textsuperscript{2},
Brucek Khailany\textsuperscript{2}
}

\affiliation{
\textsuperscript{1}Massachusetts Institute of Technology \country{}
\textsuperscript{2}NVIDIA \\
\textsuperscript{1}zhiye@mit.edu, 
 \textsuperscript{2}\{sdai, benk, bkhailany\}@nvidia.com
}

\renewcommand{\shortauthors}{Song et al.}

\begin{abstract}
Diffusion models have revolutionized video generation, becoming essential tools in creative content generation and physical simulation. Transformer-based architectures (DiTs) and classifier-free guidance (CFG) are two cornerstones of this success,  enabling strong prompt adherence and realistic video quality. Despite their versatility and superior performance, these models require intensive computation. Each video generation requires dozens of iterative steps, and CFG doubles the required compute. This inefficiency hinders broader adoption in downstream applications.

We introduce \ourwork{}, a training-free method to accelerate video generation with guidance alignment and systematic proxy selection for reuse metrics. Through rank-order correlation analysis, our technique identifies the optimal proxy for each video model, across model families and parameter scales, thereby ensuring optimal computational reuse. We achieve $1.87\times$ and $2.37\times$ speedup on Wan2.1-1.3B and Wan2.1-14B with only 0.97\% and 0.72\% drops on the VBench-2.0 benchmark. At high speedup rates, our approach maintains superior fidelity to the base model, exceeding prior state-of-the-art approaches by 5 to 10 dB in peak signal-to-noise ratio (PSNR).



\end{abstract}

\maketitle

\thispagestyle{plain}
\pagestyle{plain}

\section{Introduction}

\begin{figure}
    \centering
    \includegraphics[width=1\linewidth]{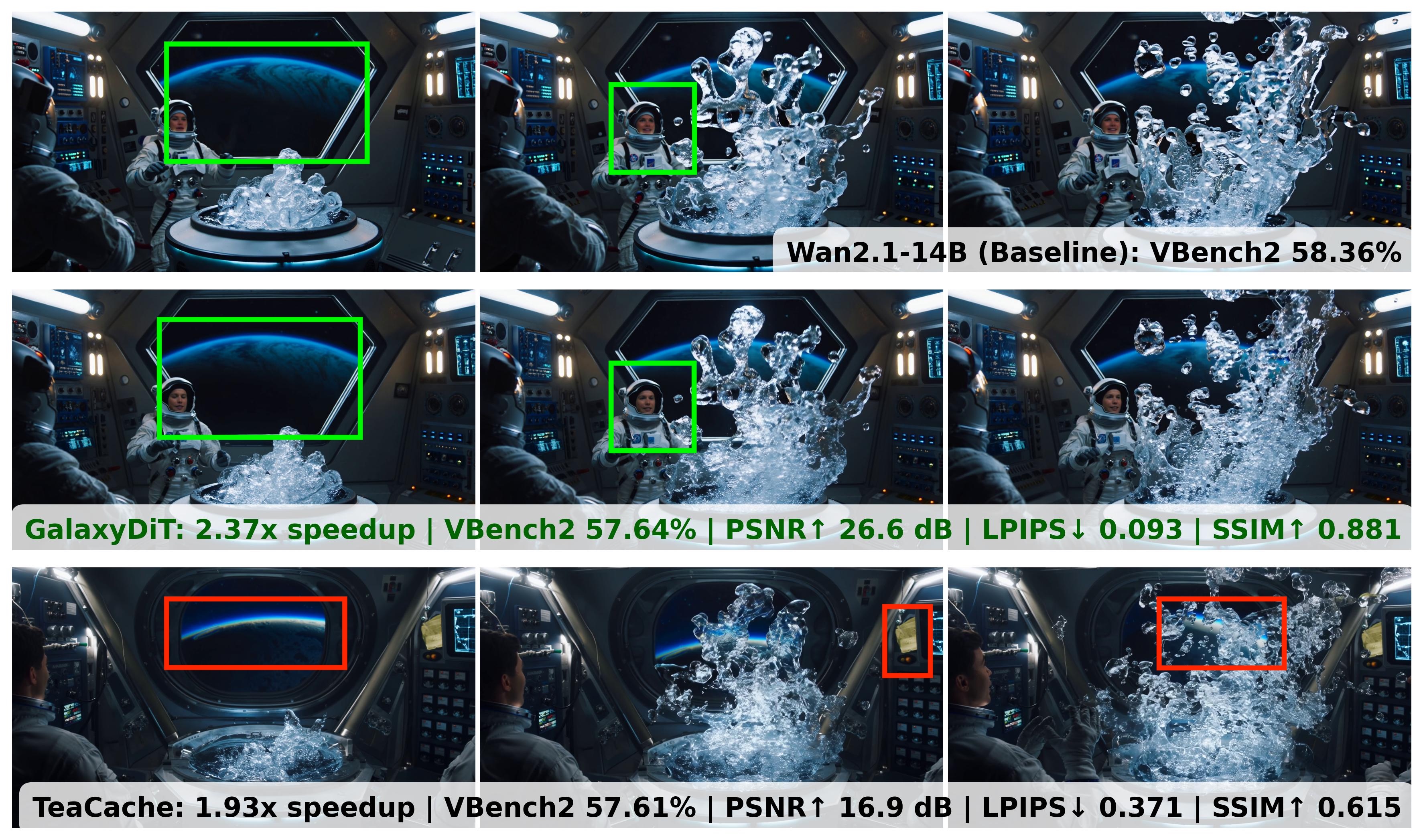}
\justifying
\footnotesize
Prompt: A space station interior with microgravity conditions, where a bottle of water is opened...
Astronauts in space suits observe the phenomenon, \textbf{their faces reflecting curiosity and amazement}. The background shows the station's control panels and windows revealing a distant planet or galaxy...
\vspace{-4pt}
\caption{Visual comparison of the Wan2.1-14B model, \ourwork{}-fast, and prior work \cite{teacache}. Metrics shown are for the full benchmark evaluation. Best viewed zoomed in.}
\vspace{-10pt}
\label{fig:fig1}
\end{figure}

Diffusion models have revolutionized image and video generation, with unprecedented capabilities in creative content production and physical world simulation \cite{image_diffusion_first,WAN,cosmos}.
Applications of video diffusion models range from entertainment and advertising to synthetic data generation for autonomous driving and robotics. Diffusion models work by learning to predict the noise that has been gradually added to videos; during inference, they progressively remove noise to generate coherent video content during iterative steps.

Diffusion transformers (DiTs), initially proposed for image generation \cite{DiT}, have become the foundation of modern video generation models \cite{WAN,hunyuanvideo,CogVideoX,cosmos,OpenSora}.
Recent state-of-the-art models, such as Wan2.1 \cite{WAN} and COSMOS \cite{cosmos}, have demonstrated the superior performance of the DiT architecture.
Classifier-free guidance (CFG) \cite{CFG} has become essential for these models \cite{hunyuanvideo,cosmos,WAN}, achieving superior prompt adherence and video quality.
These models can capture complex attributes including physical laws, commonsense reasoning, and compositional integrity, as evidenced by the VBench-2.0 benchmark \cite{vbench2}. This progression from superficial properties to modeling complex real-world phenomena makes video diffusion increasingly valuable for downstream applications.

However, the compute cost of DiT-based video generation hinders broader adoption. Each denoising step, or sampling step, requires dozens \todo{number} of Tera-FLOPs, and dozens of steps are required to produce a single video. This cost is further amplified by CFG, which runs two parallel diffusion passes at each denoising step. With these computational needs, Wan2.1-14B takes 7 minutes to generate a 5-second video clip on 8 A100 GPUs.

One promising approach to reduce computation without additional training is reuse, taking advantage of the similarities between adjacent denoising steps by saving intermediate data for later reuse. Despite feature similarities, deciding when to reuse is critical to maintain output quality. The optimal reuse opportunity differs widely across models and across prompts due to different noise schedulers, video encoders, model architectures, and parameters. Prior works have proposed a one-size-fits-all approach, using the same reuse metric for different models \cite{teacache}. However, we observe that the statistics used to calculate the reuse metric, i.e. proxies, vary across model families and scales, suggesting that a systematic approach to select the model-specific optimal reuse metric is essential.

Moreover, existing techniques treat the conditional and unconditional passes of CFG independently during reuse decisions. This misaligns the noise predictions and produces visual artifacts. Since CFG is essential for high-quality generation, developing a CFG-aware reuse strategy presents a critical opportunity for efficient video generation.

To address these challenges, we propose \ourwork{}, a training-free reuse method to accelerate video generation. Our contributions are:
\begin{itemize}
    \item We demonstrate that different models have different proxies that correlate best with the ideal reuse metric (i.e., the oracle).
    \item We design a proxy selection method using Spearman's rank correlation coefficient to identify the best proxy for each model.
    \item We propose CFG-aligned reuse to eliminate visual artifacts caused by CFG-agnostic reuse methods.
    \item We present extensive evaluations across model architectures and parameter scales. \ourwork{}-fast achieves $2.57\times$, $2.37\times$, $2.13\times$ speedups on Wan2.1-1.3B, Wan2.1-14B, and Cosmos-Predict2-2B, respectively, with high fidelity to the base model, as measured by multiple full-reference metrics. 
\end{itemize}

\begin{figure}[!tbh]
    \centering
    \includegraphics[width=1\linewidth]{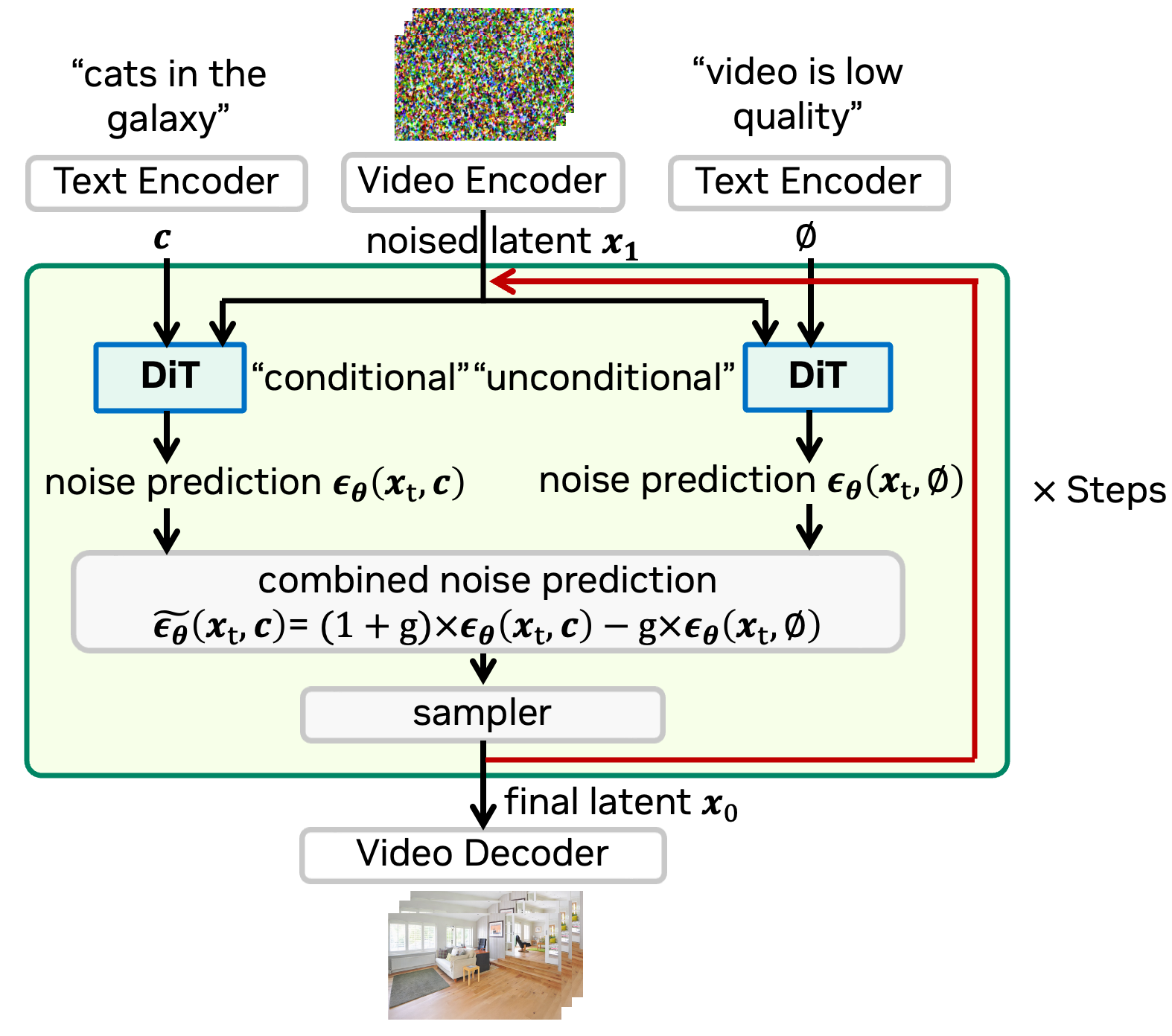}
    \caption{Video generation relies on diffusion transformers (DiTs) and classifier-free guidance (CFG).}
    \label{fig:CFG}
    \vspace{-10pt}
\end{figure}
\section{Background}
\subsection{Video Generation with DiT and CFG}
Transformer-based architecture (DiTs) and classifier-free guidance (CFG) have enabled the vast improvement in video generation \cite{video_diffusion_first,CFG}. As shown in \Cref{fig:CFG}, video generation is controlled through a condition vector $\mathbf{c}$, typically text embeddings.
Each DiT block processes the entire token sequence, through attention, cross-attention, and MLP layers. CFG \cite{CFG} runs two parallel passes, conditional (with $\mathbf{c}$) and unconditional (with $\varnothing$), then combines their noise predictions to achieve superior prompt adherence and video quality. The combined prediction is then fed to the sampler, which generates the next iteration's input. Various noise schedulers \cite{RectifiedFlow,flowmatching,image_diffusion_first} can be used to control the noise level at each denoising step.

\subsection{Related Work}

Several techniques have been proposed to reduce computation in DiT-based video generation. Model distillation approaches train student models that require fewer denoising steps.
For example, AccVideo \cite{AccVideo} trains the student model with a synthetic dataset generated by the teacher model.
MCM \cite{MCM} incorporates an additional high-quality image dataset to compensate the visual quality loss of the individual frames.
However, distillation is compute-intensive, requiring dozens of days on 8 A100 GPUs \cite{AccVideo}.

Training-free approaches avoid this cost. RADiT \cite{RADiT} skips layers when input features resemble the previous denoising step. However, it incurs significant memory cost when extended to larger models. For Wan2.1-14B, the 173 GB memory overhead makes it impossible to fit on a single A100 GPU. Similarly, the residual caching technique of DitFastAttn \cite{DiTFastAttn} saves the attention layer output, translating to 58 GB memory overhead.
FasterCache \cite{FasterCache} exploits feature similarity between CFG's conditional and unconditional passes, proposing frequency-based reuse for unconditional features, though this by design limits speedup to $2\times$.
TeaCache \cite{teacache} skips steps if the polynomial fit of the embedded input is greater than a threshold. Its extension to CFG-based models introduces visual artifacts, as will be discussed in \Cref{sec:cfg}.
Sparsity-based approaches, such as Sparse VideoGen \cite{SparseVideoGen} that uses temporal and spatial sparsity patterns, are orthogonal and can be used together with reuse-based methods. 

\subsection{Full-Reference Video Quality Metrics}

We compare latency-optimized models against base models using standard full-reference metrics: peak signal-to-noise ratio (PSNR) for absolute reconstruction error, structural similarity index measure (SSIM) for perceived structural changes, and learned perceptual image patch similarity (LPIPS) \cite{lpips} for perceptual loss.

\section{\textsc{Galaxy}DiT}

\subsection{Algorithm Overview}

\ourwork{} accelerates video generation through two key innovations: adaptive proxy selection and guidance-aligned reuse. The core insight is that not all denoising steps contribute equally to the final video quality, enabling video-dependent selective computation reuse between adjacent steps.

Our method operates in two phases: an offline proxy selection process and runtime inference acceleration. First, the proxy selection method (\Cref{sec:proxy}) identifies the optimal reuse metric for each model by analyzing rank-order correlations between candidate proxies and the oracle, which is an ideal metric of how important the current step is to the denoising process. This model-specific selection is crucial because different architectures and scales exhibit varying correlation patterns across their internal representations.

During inference, we use the selected proxy to determine when to skip computation at each denoising step. At the beginning of each step, we compute only a portion of the first DiT block of the conditional pass and extract the chosen proxy (\Cref{alg:main}, line 3). We calculate the reuse metric by comparing the current proxy to the previous step's proxy, normalized by the current proxy's magnitude (line 4). This metric accumulates across steps to track the cumulative change since the last full computation. If this accumulated metric exceeds a threshold and we are past the initial 20\% of denoising steps, we reuse cached residuals from the previous step for both conditional and unconditional passes (lines 6-8). Otherwise, we perform full computation for both passes and store the residuals for potential future reuse (lines 11-14).

A critical aspect of our approach is guidance alignment (\Cref{sec:cfg}). Unlike prior work that makes independent reuse decisions for conditional and unconditional passes, we synchronize these decisions based on the conditional pass's proxy. This ensures that the conditional and unconditional noise predictions, $\boldsymbol{\epsilon}_\theta(\mathbf{x}_t, \mathbf{c})$ and $\boldsymbol{\epsilon}_\theta(\mathbf{x}_t, \boldsymbol{\varnothing})$, used in CFG calculation (line 16), remain aligned at each denoising step, eliminating visual artifacts caused by misaligned guidance.

\begin{algorithm}
\caption{Video Generation Accelerated by \ourwork{}}
\label{alg:main}
\begin{algorithmic}[1]
\For{denoising step $S = 1$ \textbf{to} $N_{\text{steps}}$}
    \State \Comment{Run the first DiT block of the conditional generation}
    \State $\mathbf{proxy}_S \gets \text{DiTBlock0}(\mathbf{x}_t, \mathbf{c})$
    \State $\text{reuse\_metric} \gets \text{reuse\_metric} + \frac{\lVert \mathbf{proxy}_{S-1} - \mathbf{proxy}_S \rVert_1}{\lVert \mathbf{proxy}_S \rVert_1}$

    \State $\text{use-cache} \gets {S > N_{\text{steps}} \times 0.2 \,\textbf{and}\, \text{reuse\_metric} < \text{threshold}}$
    
    \If{$\text{use-cache}$}
        \State $\boldsymbol{\epsilon}_\theta(\mathbf{x}_t, \mathbf{c}) \gets \text{unpatchify}(\mathbf{y}_{\text{in},S, \text{cond}} + \mathbf{res}_{\text{cond},S-1})$
        \State $\boldsymbol{\epsilon}_\theta(\mathbf{x}_t, \boldsymbol{\varnothing}) \gets \text{unpatchify}(\mathbf{y}_{\text{in},S, \text{uncond}} + \mathbf{res}_{\text{uncond},S-1})$
    \Else
        \State $\text{reuse\_metric} \gets 0$
        \State $\boldsymbol{\epsilon}_\theta(\mathbf{x}_t, \mathbf{c}) \gets \text{DiT}(\mathbf{x}_t, \mathbf{c})$ \Comment{Run full conditional pass}
        \State $\mathbf{res}_{\text{cond},S-1} \gets \mathbf{y}_{out,S}-\mathbf{y}_{in,S}$ \Comment{occurs in the DiT}
        \State $\boldsymbol{\epsilon}_\theta(\mathbf{x}_t, \boldsymbol{\varnothing}) \gets \text{DiT}(\mathbf{x}_t, \boldsymbol{\varnothing})$ \Comment{Run full unconditional pass}
        \State $\mathbf{res}_{\text{uncond},S-1} \gets \mathbf{y}_{out,S}-\mathbf{y}_{in,S}$  \Comment{occurs in the DiT}
    \EndIf
    \State $\bar{\boldsymbol{\epsilon}}_\theta(\mathbf{x}_t, \mathbf{c}) = (1 + g) \times \boldsymbol{\epsilon}_\theta(\mathbf{x}_t, \mathbf{c}) - g \times \boldsymbol{\epsilon}_\theta(\mathbf{x}_t, \boldsymbol{\varnothing})$
    \State $\mathbf{x}_t' \gets \text{sampler}(\tilde{\boldsymbol{\epsilon}}_\theta(\mathbf{x}_t, \mathbf{c}), t)$
\EndFor
\end{algorithmic}
\end{algorithm}

\subsection{Adaptive Reuse Metric} \label{sec:proxy}

\subsubsection{The Oracle}
In an ideal scenario, we would have access to an "oracle", a metric quantifying the importance of each denoising step. We define this oracle as the relative change in residuals between consecutive steps, where the residual is the difference between the output and input of the DiT blocks, $\textbf{y}_{out}-\textbf{y}_{in}$ (\Cref{fig:proxy}). This residual quantifies the transformation applied by the DiT at each step. A larger change indicates that it is more critical for final video quality.

By comparing the oracle to a threshold, we can partition denoising steps into two sets: steps with insignificant changes that can be safely reused, and important steps requiring full computation. Quantitatively, at sampling step $S$, the oracle is:
\begin{equation}
\text{Oracle}_S=\frac{\lVert (\textbf{y}_{\text{out,S}} - \textbf{y}_{\text{in,s}}) - (\textbf{y}_{\text{out,S-1}} - \textbf{y}_{\text{in,s-1}}) \rVert_1}{\lVert (\textbf{y}_{\text{out,S}} - \textbf{y}_{\text{in,s}}) \rVert_1}
\end{equation}
where $\textbf{y}_{out}$ and $\textbf{y}_{in}$ are the output and input of the DiT blocks.

\subsubsection{Proxy Candidates}
In practice, computing the oracle requires a full forward pass through the DiT to obtain $\textbf{y}_{out}$, defeating the purpose of reuse. We therefore introduce lightweight proxies that approximate the oracle by computing only the first DiT block and collecting intermediate statistics, requiring less than 1\% of the full step computation.

Different models exhibit distinct internal dynamics during denoising. The self-attention, cross-attention, and MLP layers evolve differently across architectures and scales. Noise schedulers, which decides the amount of noise at each step, and adaptive layer norm (AdaLN) further complicate these interactions. These architectural variations mean that different intermediate tensors correlate best with step importance across models. Thus, a proxy that effectively predicts importance in one model may fail in another.

\begin{figure}[!tb]
    \centering
    \includegraphics[width=1\linewidth]{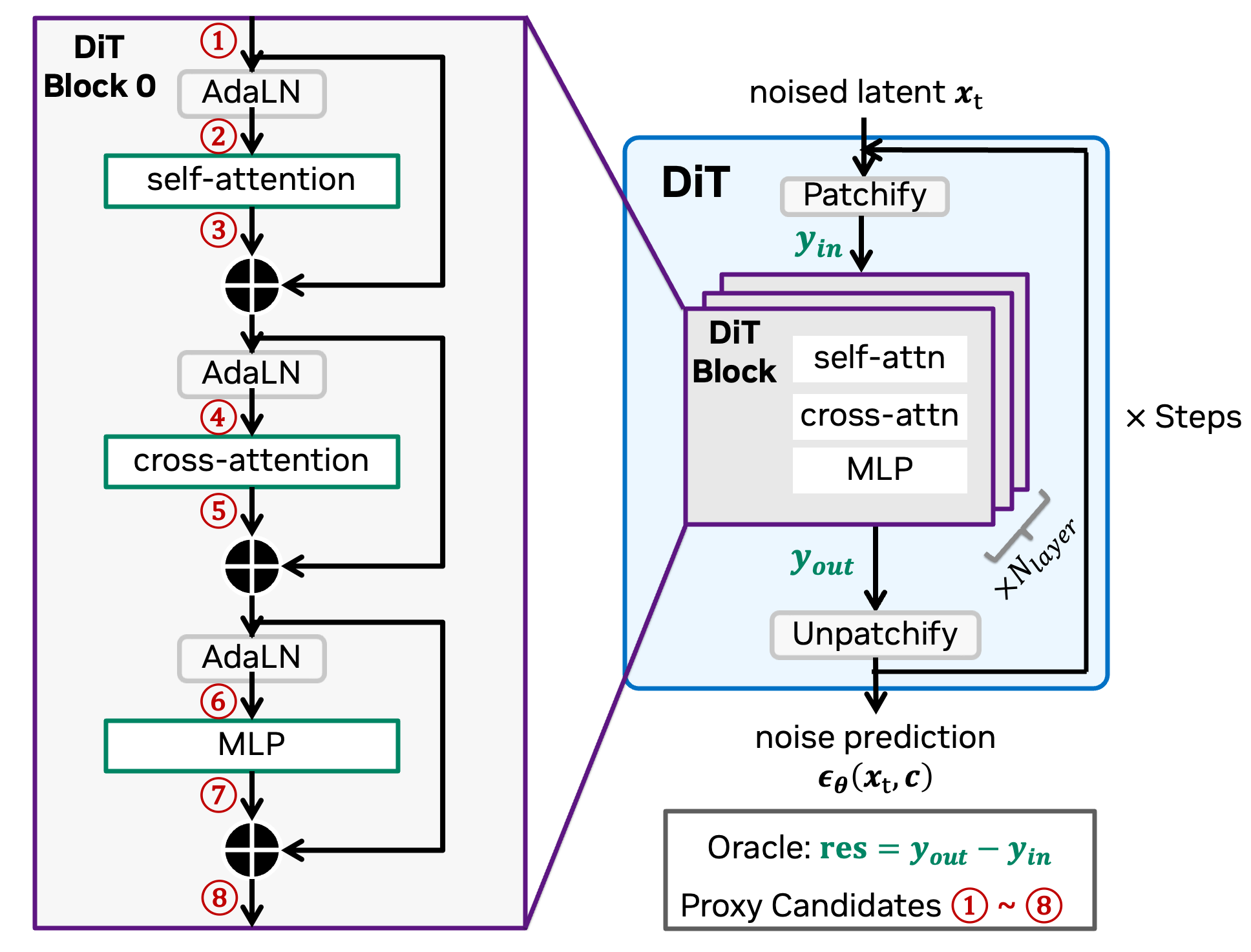}
    \caption{The oracle requires full DiT pass; various proxy candidates (names are labeled in \Cref{tab:spearman}) are intermediate tensors extracted from the first DiT block to approximate the oracle.}
    \label{fig:proxy}
\end{figure}

We propose eight proxy candidates extracted from various points in the first DiT block, as labeled in \Cref{fig:proxy}. Each proxy defines a corresponding reuse metric:
\begin{equation}
\text{Reuse metric}_S = \frac{\lVert \mathbf{proxy}_{S-1} - \mathbf{proxy}_S \rVert_1}{\lVert \mathbf{proxy}_S \rVert_1}.
\end{equation}

\subsubsection{Proxy Selection}
The key challenge is selecting the proxy that best approximates the oracle. In an ideal scenario, we would decide whether to reuse a denoising step by comparing the oracle against a threshold. In practice, we instead compare the reuse metric against another threshold. 

The reuse metric's threshold does not need to numerically match the oracle's threshold. What matters is rank-order preservation: if the oracle indicates step A is less important than step B, the reuse metric should also rank step A lower than step B. This consistent ordering ensures we skip the same set of steps regardless of the absolute metric values, as long as both the oracle and the proxy partition the steps at similar positions in their respective rankings.

To quantify order preservation, we use Spearman's rank correlation coefficient $\rho$, which evaluates the order-preserving nature of two data sets. Spearman's $\rho$ is equivalent to the Pearson correlation of the rank values of two variables:
\begin{equation}
\rho = 1 - \frac{6 \sum_{i=1}^{n} d_i^2}{n(n^2 - 1)},
\end{equation}
where $d_i = \operatorname{rank}(x_i) - \operatorname{rank}(y_i)$ represents the rank disagreement between corresponding elements of the two datasets.

In our context, we compute ranks across the denoising steps of a single video generation. Specifically, for each step $i \in \{1, \ldots, N_{\text{steps}}\}$:
\begin{equation}
d_i = \operatorname{rank}(\text{oracle}_i) - \operatorname{rank}(\text{reuse metric}_i).
\end{equation}
A high Spearman's $\rho$ (close to 1) indicates that the proxy candidate preserves the oracle's ordering, making it suitable for our method. We compute $\rho$ for each proxy candidate across a small number of prompts and select the proxy with the highest average correlation as the optimal choice for that model.

\subsubsection{Proxy Selection Results}

\begin{figure}
    \centering
    \begin{subfigure}[b]{0.46\linewidth}
        \centering
        \includegraphics[width=\linewidth]{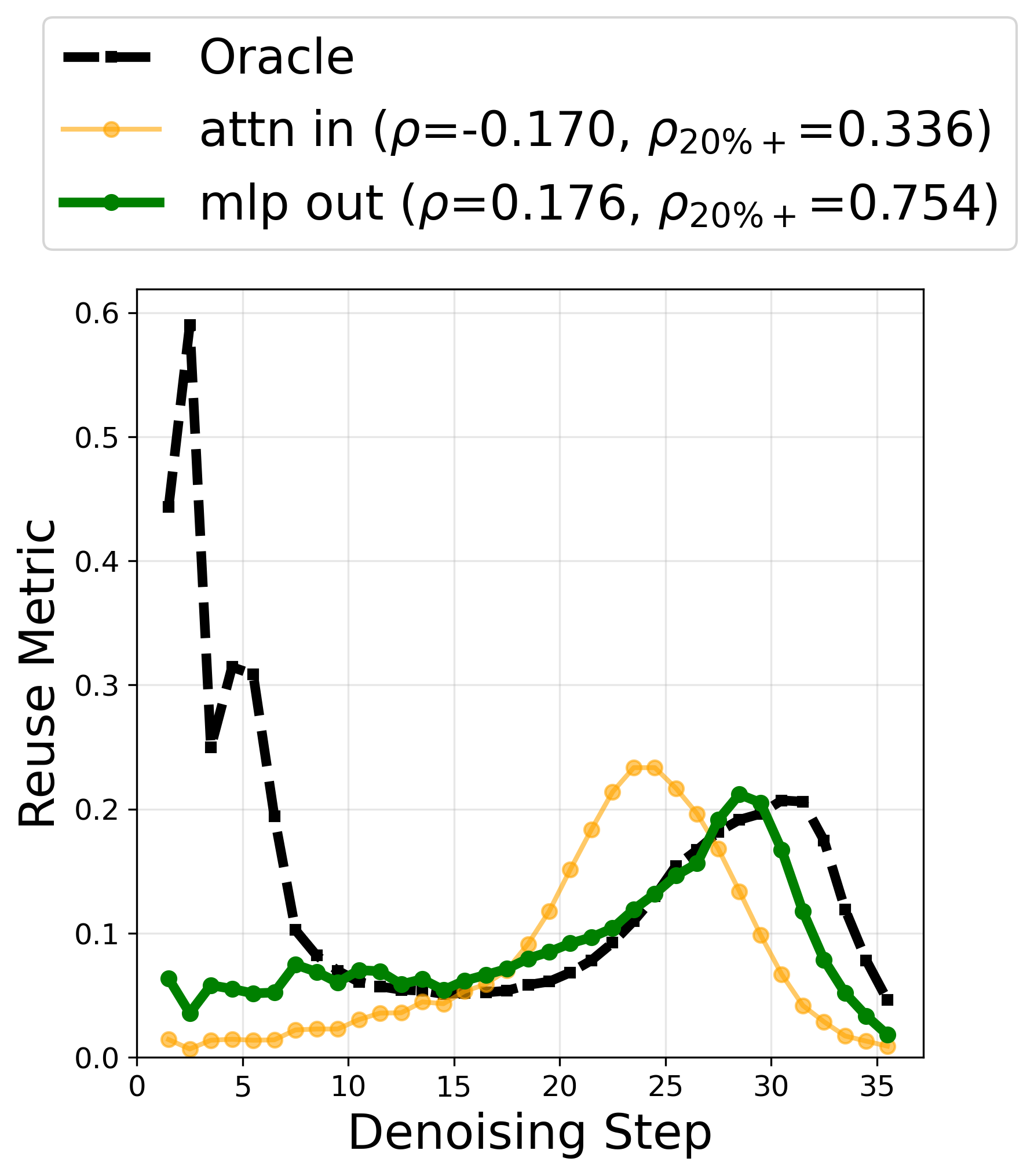}
        \caption{Cosmos-Predict2-2B}
        \label{fig:p2-proxy-a}
    \end{subfigure}
    \hfill
    \begin{subfigure}[b]{0.52\linewidth}
        \centering
        \includegraphics[width=\linewidth]{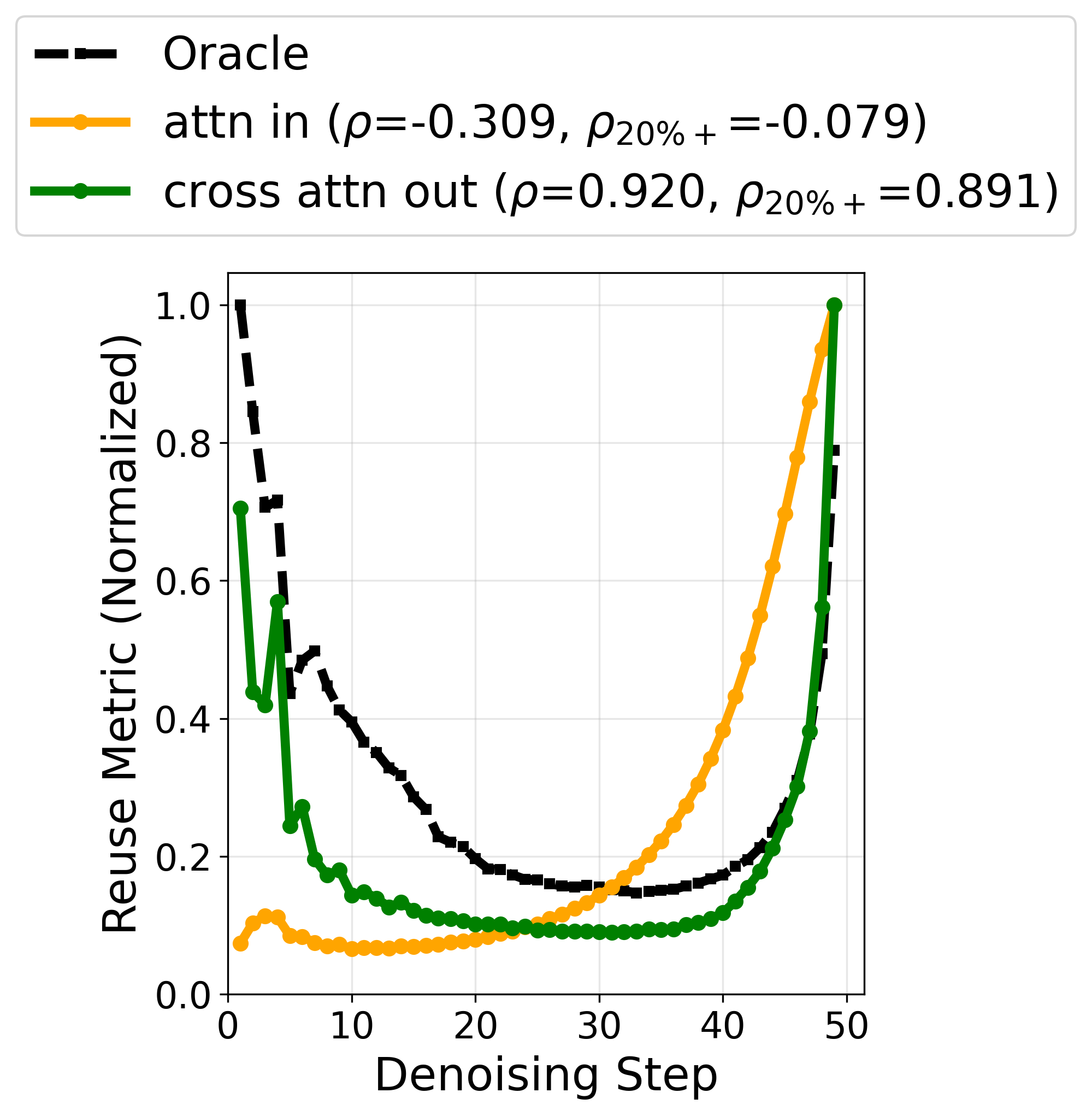}
        \caption{Wan2.1-1.3B}
        \label{fig:p2-proxy-b}
    \end{subfigure}
    \caption{The comparison of two proxy candidates against the oracle for an example prompt. The optimal proxy candidate for each model is shown in green. A suboptimal proxy candidate is shown in orange.}
    \label{fig:p2-proxy}
    \vspace{-5pt}

\end{figure}

\Cref{fig:p2-proxy} visualizes how different proxy candidates compare to the oracle in two model families. For Cosmos-Predict2 (\Cref{fig:p2-proxy-a}), \texttt{mlp out} coincides closely with the oracle, while \texttt{attn in} peaks earlier, failing to capture the oracle's behavior. Due to the lack of proxies reflecting the oracle at early steps, we exclude the first 20\% of steps from reuse consideration, improving Spearman's $\rho$ from 0.176 to 0.754. This exclusion strategy is adopted across all models. For Wan2.1-1.3B (\Cref{fig:p2-proxy-b}), \texttt{cross attn out} serves as the best oracle indicator.

Our analysis reveals two critical findings. First, the optimal proxy exhibits consistency across prompts within the same model. For Wan2.1-1.3B, $\rho_{\text{cross attn out}}$ achieves a mean of 0.89 with a standard deviation of only 0.11, demonstrating strong prompt-independent correlation. Second, as shown in \Cref{tab:spearman}, the three models have three different optimal proxies. These results underscore the necessity of model-specific proxy selection. Using a suboptimal proxy can severely degrade performance. For instance, applying the \texttt{mlp out} (optimal for Predict2) to Wan2.1-14B yields a weak correlation of only $\rho = 0.21$, making it unsuitable for reliable reuse decisions.
\begin{table}[!tb]
\centering
\small
\setlength{\tabcolsep}{2.5pt}
\caption{Spearman's correlation ($\rho$) between proxy candidates and oracle. Optimal proxies are bolded.}
\label{tab:spearman}
\begin{tabular}{lccc}
\hline
Proxy & Cosmos-Predict2-2B & Wan2.1-1.3B & Wan2.1-14B \\
\hline
\textcircled{1} block in & 0.16 & 0.35 & 0.00 \\
\textcircled{2} attn in & 0.45 & 0.35 & 0.02 \\
\textcircled{3} attn out & -0.07 & 0.60 & \textbf{0.65} \\
\textcircled{4} cross attn in & 0.32 & 0.35 & 0.03 \\
\textcircled{5} cross attn out & 0.68 & \textbf{0.89} & 0.53 \\
\textcircled{6} mlp in & 0.64 & 0.40 & 0.16 \\
\textcircled{7} mlp out & \textbf{0.82} & 0.40 & 0.21 \\
\textcircled{8} block out & 0.51 & 0.40 & 0.13 \\
\hline
\end{tabular}
\vspace{0pt}
\end{table}

\begin{figure}[!tb]
    \centering
    \begin{subfigure}[b]{\linewidth}
        \centering
        \includegraphics[width=\linewidth]{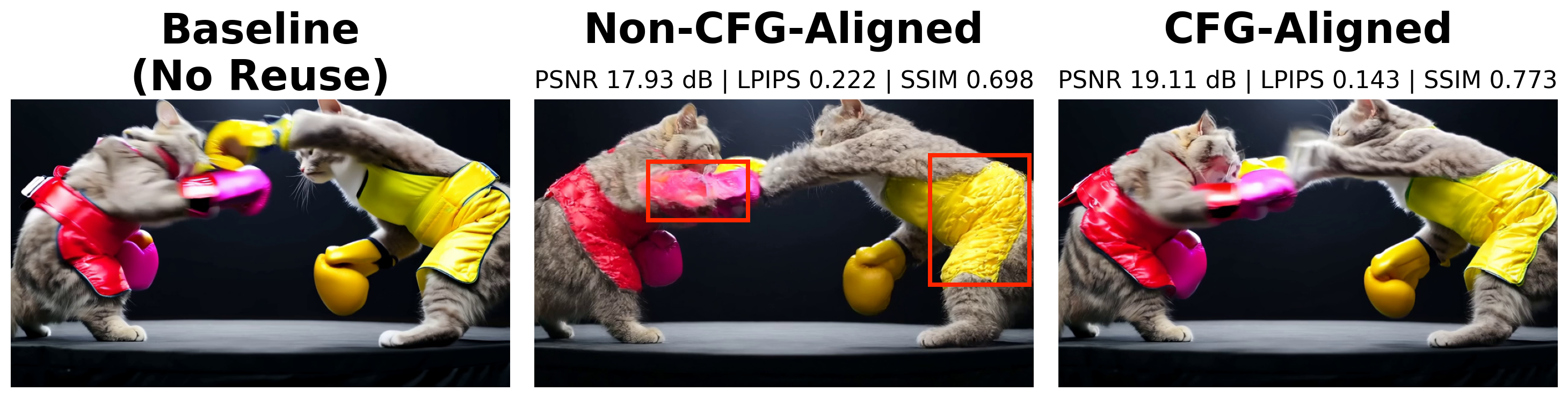}
        \caption{Wan2.1-1.3B}
        \label{fig:mtv:1.3b}
    \end{subfigure}
    
    \vspace{0pt}
    
    \begin{subfigure}[b]{\linewidth}
        \centering
        \includegraphics[width=\linewidth]{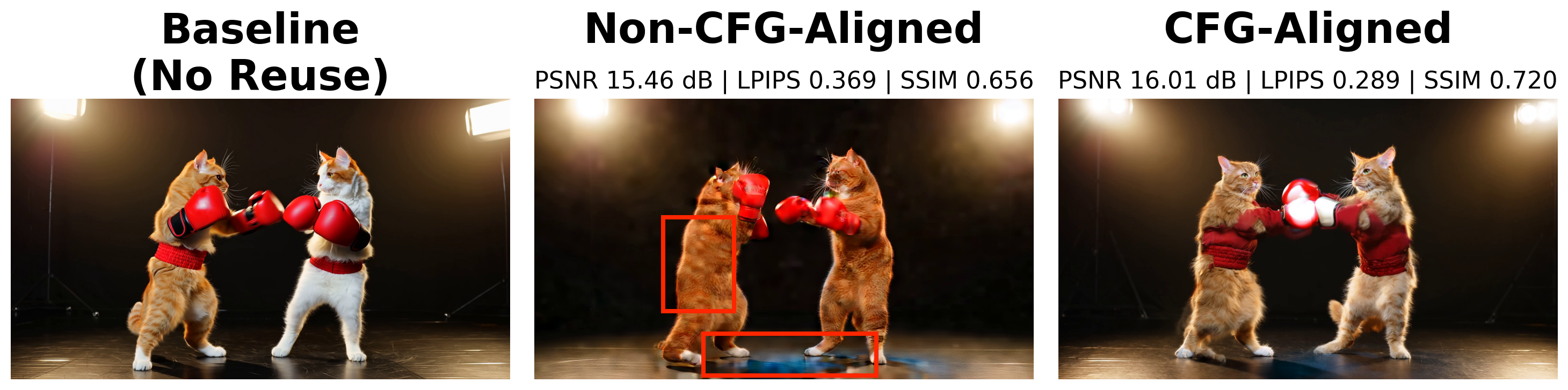}
        \caption{Wan2.1-14B}
        \label{fig:mtv:14b}
    \vspace{-8pt}
    \end{subfigure}
    
    \caption{Impact of CFG alignment on visual quality. Visual artifacts in the non-CFG-aligned video is highlighted. Best viewed zoomed in.}
    \label{fig:mtv}
    \vspace{-8pt}
\end{figure}

\subsection{Guidance-Aligned Reuse} \label{sec:cfg}

Prior work typically determines step skipping independently for the conditional and unconditional passes \cite{teacache}. However, this independence causes visual artifacts. When the conditional DiT is computed at the current step but the unconditional noise prediction is reused from a previous step, the two predictions come from slightly different noise levels. Since CFG combines these predictions (\Cref{alg:main}, line 16), this mismatch fails to steer the diffusion process from low-quality outputs, producing visual artifacts.

\Cref{fig:mtv} illustrates this phenomenon at a high reuse ratio to amplify visual differences, though the effect persists at lower ratios (quantified in \Cref{sec:ablation_cfg}). In the non-aligned approach, the clothing in \Cref{fig:mtv:1.3b} appears less vibrant, while the fur and floor in \Cref{fig:mtv:14b} show blurriness and anomalous artifacts. In contrast, both the baseline and CFG-aligned versions preserve fine details such as fur texture.

To address this, we propose guidance-aligned reuse: the proxy from the first DiT block of the conditional pass (line 3 of \Cref{alg:main}) determines whether to reuse at each step. Crucially, both conditional and unconditional passes then either compute or reuse together (lines 7\&8 and 11\&13). This synchronization ensures the guided noise prediction in line 16 receives matching inputs from both passes, eliminating visual artifacts caused by misalignment.

\section{Evaluation}
\subsection{Experimental Setup}

We evaluate \ourwork{} on three state-of-the-art DiT-based video diffusion models representing different scales, architectures, and modalities. Wan2.1-1.3B and Wan2.1-14B \cite{WAN} are leading text-to-video (T2V) models widely used in artistic content generation. Cosmos-Predict2-2B \cite{nvidia_cosmos-predict2_2025} is a foundation model targeting physical world simulation. It is an image-to-video (I2V) model, which generates a video from an initial frame and a text prompt. Wan2.1 models use 50 denoising steps while Cosmos-Predict2 uses 36 steps. For each model, we select the optimal proxy using statistics collected on 16 prompts as described in \Cref{sec:proxy}, then determine representative thresholds for different quality-latency trade-offs.

We evaluate Wan2.1 models on all 1330 prompts from the VBench-2.0 suite~\cite{vbench2}, and Cosmos-Predict2 on the 1118 image-prompt pairs from the VBench(beta)-I2V dataset~\cite{vbench++,huang_vbench-i2v_2024}. Latency is measured on 1 A100 GPU for Wan2.1-1.3B and Cosmos-Predict2, and 8 A100s for Wan2.1-14B.

The baseline is the original model~\cite{WAN,cosmos} with full computation. Baseline-generated videos serve as references for computing PSNR, SSIM, and LPIPS. VBench-2.0 scores are computed using the official implementation.
\subsection{Ablation on Reuse Metric Selection} \label{sec:ablation_metric}

\begin{figure}
    \centering
    \includegraphics[width=1\linewidth]{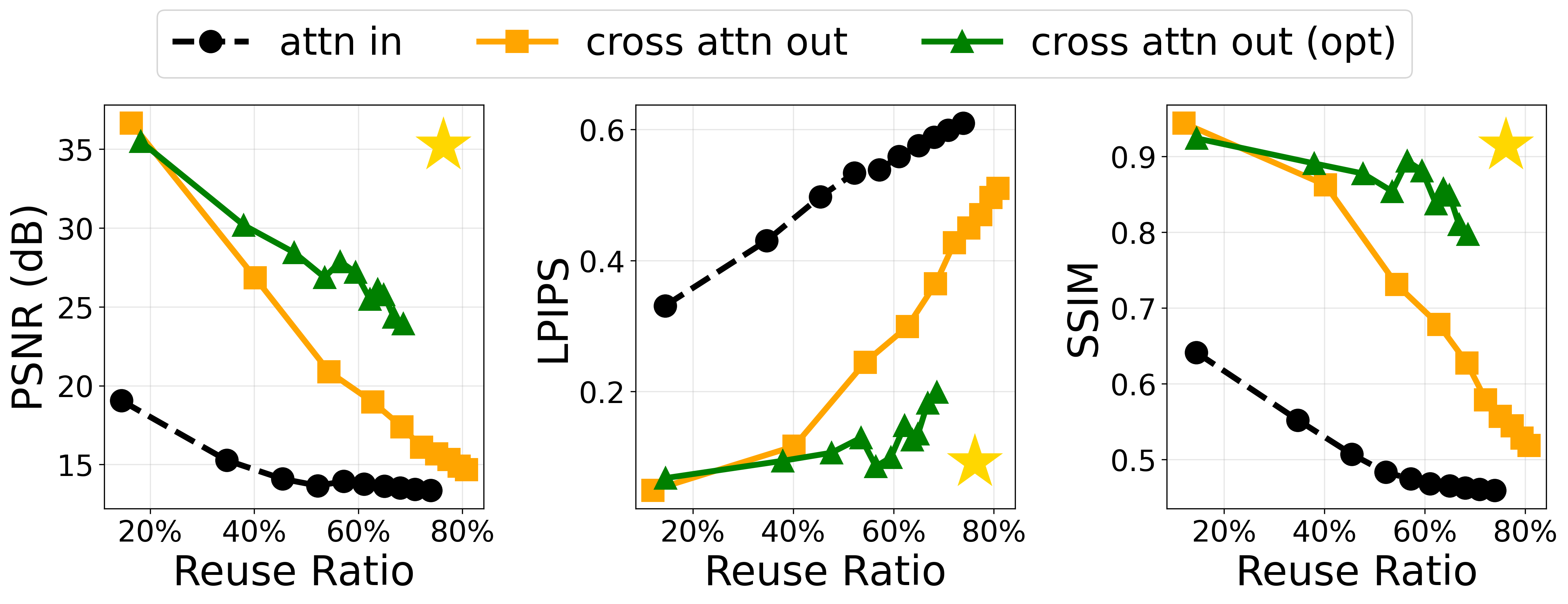}
    \caption{Ablation of proxy selection for Wan2.1-1.3B. \texttt{Cross attn out (opt)} uses \texttt{cross attn out} as the proxy and excludes the first 20\% of denoising steps from reuse. The star marks the optimal corner with higher quality and reuse.}
    \vspace{-10pt}
    \label{fig:ablation_metric}
\end{figure}

We ablate proxy selection on Wan2.1-1.3B using a random subset of VBench-2.0 prompts with CFG-aligned mode. We compare three configurations: \texttt{attn in}, \texttt{cross attn out}, and \texttt{cross attn out (opt)}. The purpose of this ablation is to conclude if our Spearman's-coefficient-based selection criterion leads to higher video quality.

Based on Spearman's coefficients in \Cref{tab:spearman}, \texttt{attn in} correlates weakly with the oracle ($\rho=0.35$), while \texttt{cross attn out} shows strong order-preservation ($\rho=0.89$). \Cref{fig:ablation_metric} confirms that the video quality using the optimal proxy is higher: \texttt{cross attn out} achieves up to 15 dB higher PSNR at lower reuse ratios, with further improvement at high reuse when excluding initial steps.

\subsection{Ablation on CFG-Alignment}\label{sec:ablation_cfg}
Using the optimal reuse metric identified in \Cref{sec:ablation_metric}, \texttt{cross attn out (opt)}, we compare CFG-aligned versus non-aligned approaches on the same VBench-2.0 subset for Wan2.1-1.3B.

The conventional approach determines reuse independently for conditional and unconditional passes at each step. Our CFG-aligned approach synchronizes these decisions: reuse always occurs simultaneously for both passes.

\begin{figure}
    \centering
    \includegraphics[width=1\linewidth]{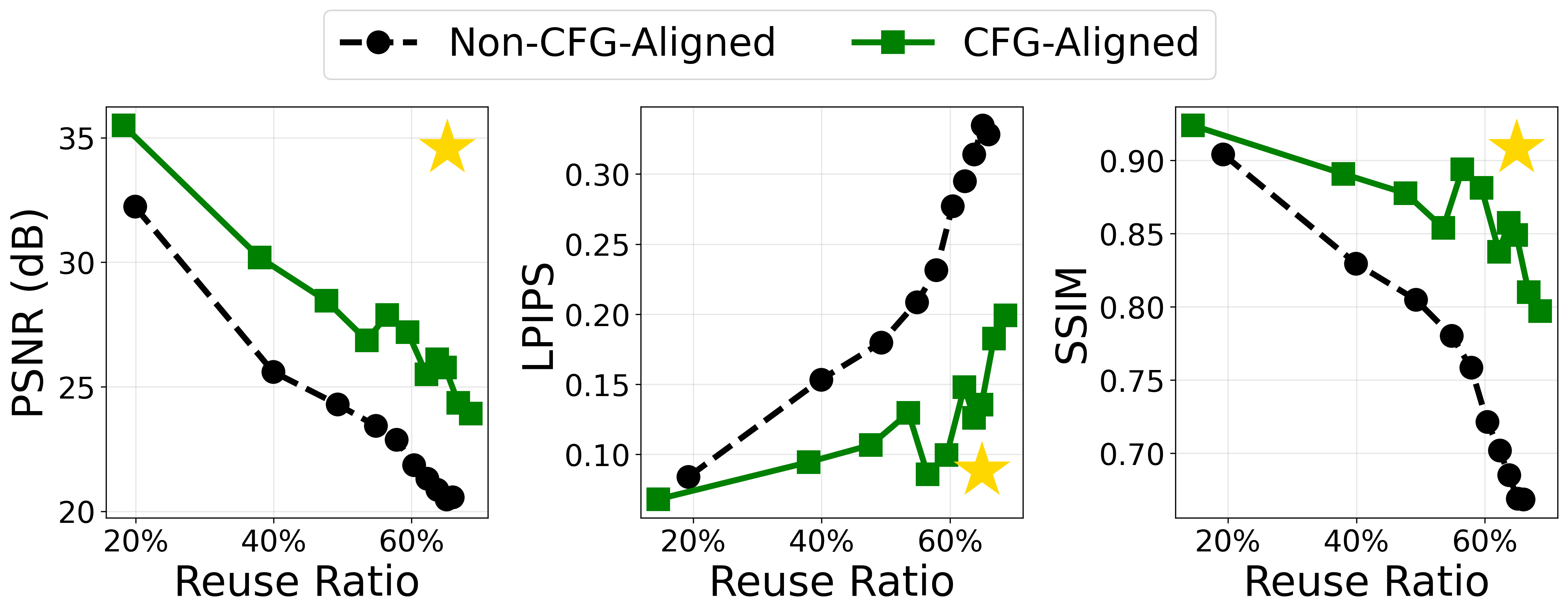}
    
    \caption{Ablation of guidance alignment in Wan2.1-1.3B. Alignment improves all quality metrics across reuse rates.
  }
    \label{fig:ablation_cfg}
\end{figure}

As shown in \Cref{fig:ablation_cfg}, CFG alignment achieves 4 dB higher PSNR, 10\% lower perceptual loss, and 10\% higher structural similarity at 60\% reuse, confirming the visual motivation in \Cref{fig:mtv}.

\subsection{Quality and Speedup}

\begin{figure}[!tb]
    \centering
    \begin{subfigure}[b]{\linewidth}
        \centering
        \includegraphics[width=\linewidth]{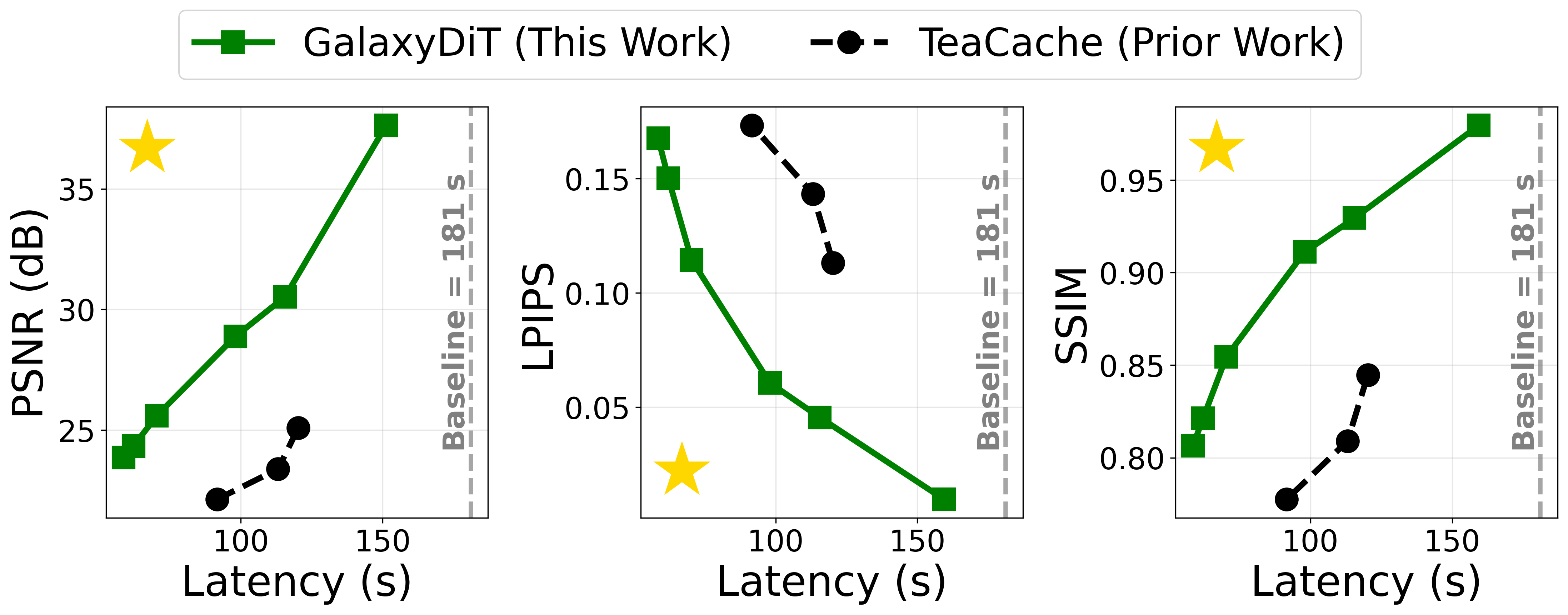}
\caption{Wan2.1-1.3B \textnormal{(T2V, 81 frames, $832\times480$)}}
        \label{fig:eval:1.3b}
    \end{subfigure}
    \vspace{0cm}
    \begin{subfigure}[b]{\linewidth}
        \centering
        \includegraphics[width=\linewidth]{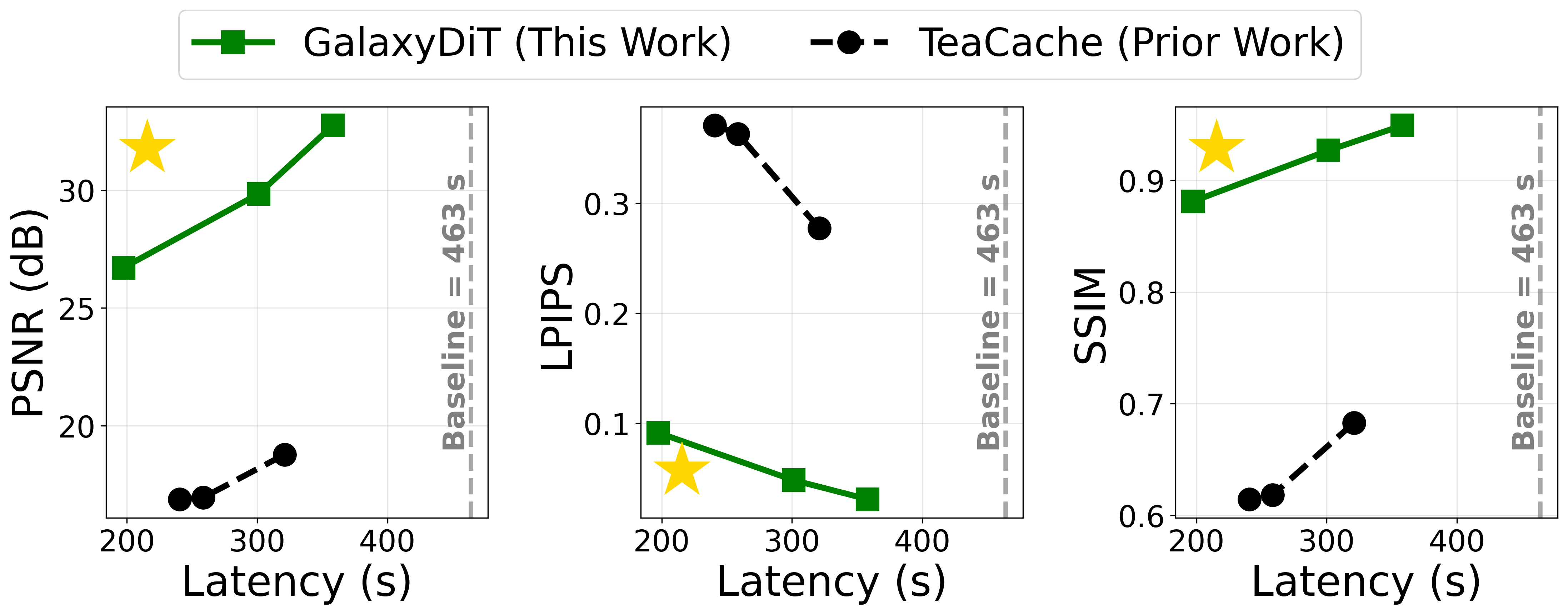}
\caption{Wan2.1-14B \textnormal{(T2V, 81 frames, $1280\times720$)}}
        \label{fig:eval:14b}
    \end{subfigure}
        \vspace{0cm}
    \begin{subfigure}[b]{\linewidth}
        \centering
        \includegraphics[width=\linewidth]{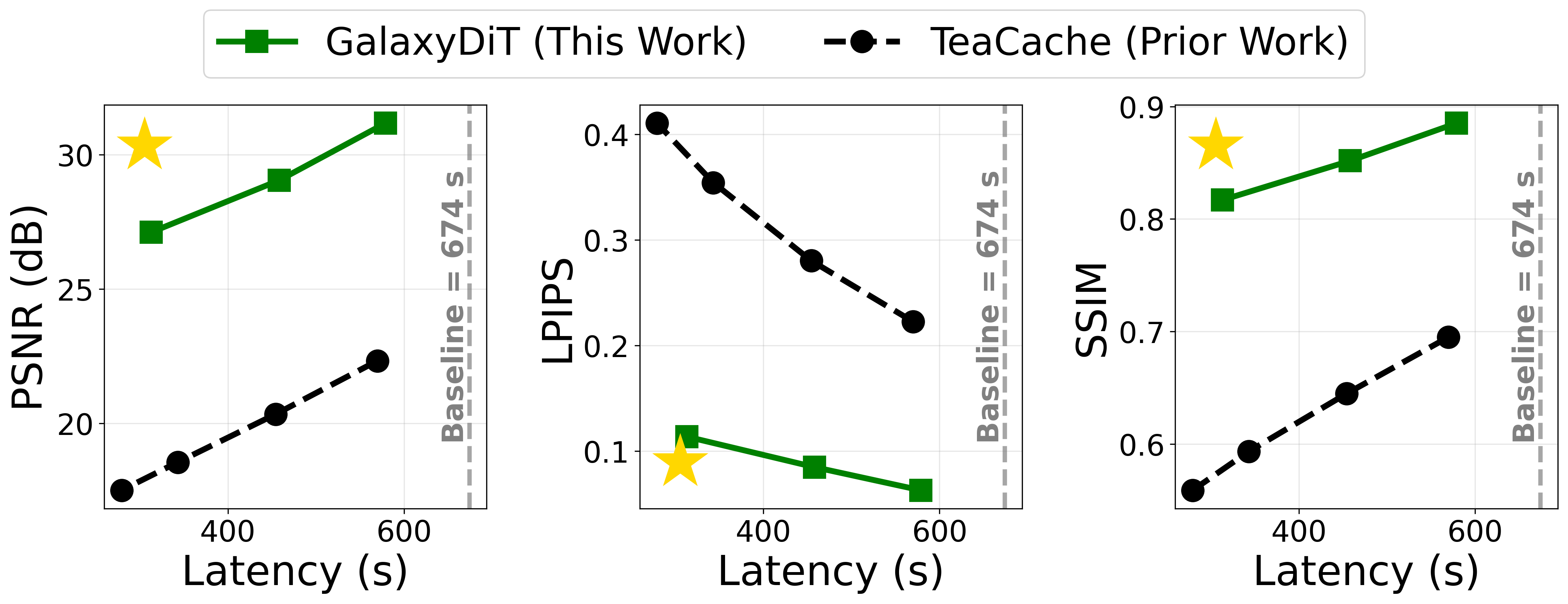}
\caption{Cosmos-Predict2-2B \textnormal{(I2V, 93 frames, $1280\times704$)}}
        \label{fig:eval:cosmos}
    \end{subfigure}
    \caption{Quality-latency comparison of \ourwork{} against prior work \cite{teacache} and the full-compute baseline. \ourwork{} achieves superior quality at equivalent speedup across T2V and I2V modalities. The yellow star indicates the optimal corner.}
    \label{fig:eval}
\end{figure}

Selecting different thresholds trades off video quality for speed. \ourwork{}-fast uses higher thresholds for maximum speedup, while \ourwork{}-slow prioritizes quality, which is suitable for quality-sensitive applications like robotics simulation.

At matched quality (25.6 dB PSNR, 0.11 LPIPS, and 0.85 SSIM), \ourwork{} achieves $1.74\times$ speedup over prior work and $2.57\times$ over baseline on Wan2.1-1.3B (\Cref{fig:eval:1.3b}). At matched latency, \ourwork{} delivers 6 dB higher PSNR, 10\% lower perceptual loss, and 10\% higher structural similarity. This trend holds across scales: on Wan2.1-14B, \ourwork{} maintains above 26.6 dB PSNR, 88\% SSIM, and below 9.3\% LPIPS at $2.37\times$ speedup. This trend also holds across model families: on Cosmos-Predict2, at the same speedup, \ourwork{} achieves 9 dB higher PSNR, 10\% lower LPIPS and 20\% higher SSIM.

We evaluate on VBench-2.0 rather than VBench-1.0 because recent models have saturated VBench-1.0's superficial metrics, and emerging applications require the intrinsic faithfulness attributes, including physical law adherence, commonsense reasoning, and compositional integrity, which are measured by VBench-2.0 \cite{vbench2}. As tested on the VBench-2.0 prompts, even \ourwork{}-fast exceeds the fidelity of prior work's slow variant while delivering $2.57\times$ (1.3B) and $2.37\times$ (14B) speedups (\Cref{tab:vbench}). Notably, \ourwork{}-slow improves the base 14B model's VBench-2.0 score by $+0.8\%$.

\begin{table}[!tb]
\centering
\small
\setlength{\tabcolsep}{2pt}
\begin{tabular}{lccccc}
\toprule
Method & Speedup$\uparrow$ & PSNR$\uparrow$ & LPIPS$\downarrow$ & SSIM$\uparrow$ & VBench-2.0$\uparrow$ \\
\midrule
\multicolumn{6}{c}{\textbf{Wan2.1-1.3B} (81 frames, $832\times480$)} \\
\midrule
Baseline & $1.00\times$ & -- & -- & -- & \textbf{56.65\%} \\
TeaCache-slow  & $1.51\times$ & 25.1 dB & 0.113& 0.845& 55.39\% ($-1.26\%$)\\
TeaCache-fast  & $1.98\times$ & 22.1 dB & 0.173& 0.777& 52.91\% ($-3.74\%$)\\
GalaxyDiT-slow & $1.85\times$& \textbf{28.9 dB} & \textbf{0.061}& \textbf{0.911}& 55.68\% ($-0.97\%$)\\
GalaxyDiT-fast & $\boldsymbol{2.57\times}$ & 25.6 dB & 0.114& 0.854& 52.83\% ($-3.82\%$)\\
\midrule
\midrule
\multicolumn{6}{c}{\textbf{Wan2.1-14B} (81 frames, $1280\times720$)} \\
\midrule
Baseline & $1.00\times$ & -- & -- & -- & 58.36\% \\
TeaCache-slow  & $1.45\times$ & 18.8 dB & 0.277& 0.683& 58.79\% ($+0.43\%$)\\
TeaCache-fast  & $1.93\times$ & 16.9 dB & 0.371& 0.615& 57.61\% ($-0.75\%$)\\
GalaxyDiT-slow & $1.30\times$ & \textbf{32.7 dB} & \textbf{0.031}& \textbf{0.949}& \textbf{59.22\%} ($+0.86\%$)\\
GalaxyDiT-medium& $1.54\times$ & 29.8 dB & 0.048& 0.927& 58.05\% ($-0.31\%$)\\
GalaxyDiT-fast & $\boldsymbol{2.37\times}$& 26.6 dB & 0.093& 0.881& 57.64\% ($-0.72\%$)\\
\bottomrule
\end{tabular}
\caption{Performance comparison of GalaxyDiT against prior works on Wan2.1 models.}
\label{tab:vbench}
\vspace{-10pt}
\end{table}

\subsection{Visual Comparison}
\begin{figure}
    \centering
    \includegraphics[width=1\linewidth]{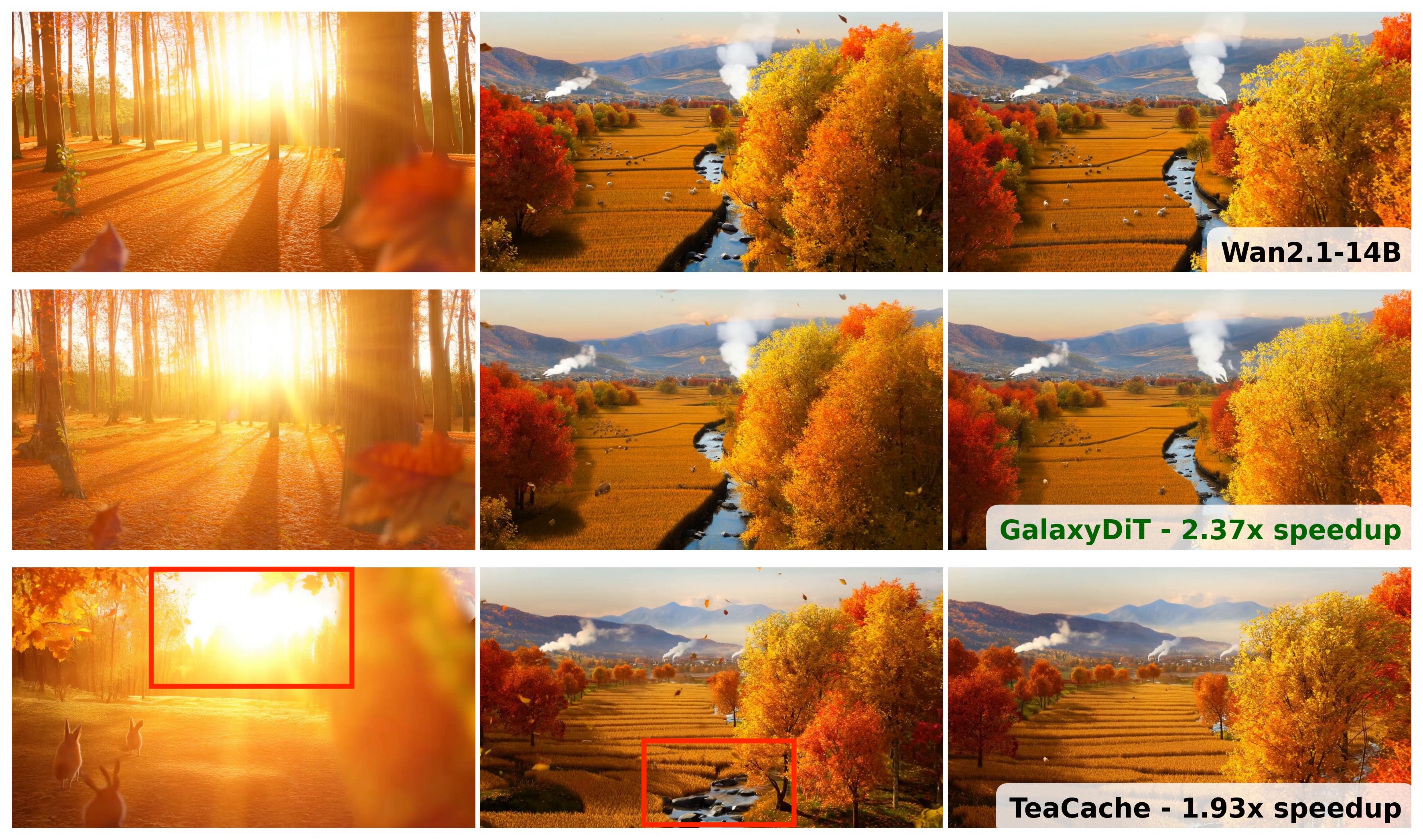}
\justifying
\footnotesize
Prompt: The camera enters a golden autumn forest, where the leaves have turned brilliant shades from gold to orange-red. A few leaves drift down with the wind. \textbf{Sunlight filters through the gaps in the trees}, casting dappled spots of light on the forest floor. \textbf{The scene shifts} to a field outside the forest, where ripe rice stalks sway in the breeze, their golden heads bowing under the weight of the grain...
The camera moves again to \textbf{a winding stream}...
\vspace{-5pt}
\caption{Qualitative comparison on a complex landscape scene with Wan2.1-14B. Best viewed zoomed in.}
    \label{fig:eg1}
\vspace{-10pt}
\end{figure}

\Cref{fig:fig1} and \Cref{fig:eg1} show example videos from VBench-2.0 prompts. In both scenarios, \ourwork{} closely resembles the baseline. The missing planetary details in TeaCache's version (\Cref{fig:fig1}) contribute to its low PSNR, since PSNR measures pixel-level fidelity. Similarly in \Cref{fig:eg1}, tree details basked in the sunlight are preserved in \ourwork{} but lost in prior work. SSIM captures structural similarity, and both examples show that the structural composition of the prior work is different from the base model, despite using the same noised latent.

Beyond fidelity, TeaCache exhibits worse prompt adherence. The space station prompt specifies astronauts with faces "reflecting curiosity and amazement," yet TeaCache omits one astronaut entirely while the other faces away from camera. This degradation likely stems from the non-CFG-aligned method, given that CFG is essential for prompt adherence.

\subsection{Memory Efficiency}
We analyze memory overhead of the 14B model as it represents the most memory-constrained scenario. The model itself contains 28 GB of parameters, with the text encoder requiring an additional 20 GB that can be offloaded during diffusion \cite{WAN}. 

\ourwork{} stores four tensors: residuals from the previous step ($\mathbf{res}_{S-1,cond}$ and $\mathbf{res}_{S-1,uncond}$), the current pass input for residual calculation ($\mathbf{y}_{in,S,\text{current pass}}$), and the conditional proxy for computing the next reuse metric ($\mathbf{proxy}_{S,cond}$). For a sequence length of 75600, this amounts to 2.88 GB of GPU memory.

For single-GPU inference (e.g., A100 40 GB or 80 GB), memory is critical. RADiT \cite{RADiT} stores tensors at every layer (173 GB), making it prohibitively expensive, while TeaCache \cite{teacache} requires 3.6 GB for embedded inputs and cached tensors. \ourwork{} achieves the lowest memory overhead among reuse-based acceleration methods.

\section{Conclusion}
We presented \ourwork{}, a training-free method to accelerate video diffusion transformers through guidance alignment and adaptive proxy-based reuse. By systematically selecting optimal proxies for each model using rank-order correlation analysis, we ensure that reuse decisions align with the true importance of each denoising step. Our guidance-aligned reuse strategy eliminates visual artifacts inherent in CFG-agnostic approaches. Evaluated on Wan2.1-1.3B and Wan2.1-14B, GalaxyDiT achieves $1.87\times$ and $2.37\times$ speedup with only 0.97\% and 0.72\% drop on the VBench-2.0 benchmark, respectively. With both the T2V Wan2.1 models and the I2V Cosmos-Predict2, \ourwork{} achieves high fidelity to the base model, 5 to 10 dB higher PSNR than prior works at equivalent speedups. With model-specific proxy selection and CFG-aware reuse, \ourwork{} enables downstream applications with efficient and high-quality video generation.

\clearpage
\newpage
\newpage
\begin{acks}
The authors would like to thank Weili Nie, Reena Elangovan, Marina Neseem, Charbel Sakr, Celine Lin, Rangharajan Venkatesan, Xiao Fu, Akshat Ramachandran, Jianming Tong, Jacob Ridgway and Bryan Foo for insightful discussions.

\end{acks}

\end{document}